\documentclass[10pt, a4paper]{article}
\usepackage{lrec2022} 
\usepackage{multibib}
\newcites{languageresource}{Language Resources}
\usepackage{graphicx}
\usepackage{tabularx}
\usepackage{soul}
\usepackage{titlesec}
\titleformat{\section}{\normalfont\large\bfseries\center}{\thesection.}{1em}{}
\titleformat{\subsection}{\normalfont\SmallTitleFont\bfseries\raggedright}{\thesubsection.}{1em}{}
\titleformat{\subsubsection}{\normalfont\normalsize\bfseries\raggedright}{\thesubsubsection.}{1em}{}
\renewcommand\thesection{\arabic{section}}
\renewcommand\thesubsection{\thesection.\arabic{subsection}}
\renewcommand\thesubsubsection{\thesubsection.\arabic{subsubsection}}

\usepackage{epstopdf}
\usepackage[utf8]{inputenc}

\PassOptionsToPackage{hyphens}{url}\usepackage{hyperref} 
\usepackage{xstring}

\usepackage{color}

\usepackage{todonotes}

\makeatletter
\def\new@fontshape{}
\makeatother

\usepackage{siunitx}
\usepackage{tipa}
\usepackage{gb4e}

\usepackage{xspace}
\newcommand{\tugebic}{{\sc t}u{\sc g}e{\sc b}i{\sc c}\xspace}

\title{TuGeBiC -- A Turkish German Bilingual Code-Switching Corpus}

\name{Jeanine Treffers-Daller^{1}, Özlem Çetino\u{g}lu^{2}} 

\address{{}^{1}Department of English Language and Applied Linguistics, University of Reading\\
         {}^{2}Institute for Natural Language Processing (IMS), University of Stuttgart \\
         j.c.treffers-daller@reading.ac.uk, ozlem.cetinoglu@ims.uni-stuttgart.de \\
        }
    
\abstract{
In this paper we describe the process of collection, transcription, and annotation of recordings of spontaneous speech samples from Turkish-German bilinguals, and the compilation of a corpus called \tugebic. Participants in the study were adult Turkish-German bilinguals living in Germany or Turkey at the time of recording in the first half of the 1990s. The data were manually tokenised and normalised, and all proper names (names of participants and places mentioned in the conversations) were replaced with pseudonyms. Token-level automatic language identification was performed, which made it possible to establish the proportions of words from each language. The corpus is roughly balanced between both languages. We also present quantitative information about the number of code-switches, and give examples of different types of code-switching found in the data. The resulting corpus has been made freely available to the research community.\\ 
}

\begin{document}

\maketitleabstract

\section{Introduction}

In many bilingual speech communities around the world, bilinguals switch between languages within one conversation and even within one utterance. This behaviour, which is generally called code-switching\footnote{Some researchers prefer the term  \textsc{translanguaging} \cite{creese2010translanguaging} for pedagogical practices which allow teachers and learners to make use of all their language resources in classroom settings. As only a few transcripts in the TuGeBic data were recorded in classrooms, the term translanguaging is less appropriate for our purposes.} presents a challenge for linguists trying to explain where switching is possible within one sentence, as well as for psychologists interested in bilingual speech processing. For linguists, it is very important to obtain information about the variability in code-switching patterns: it has become clear that there is a wide range of different code-switching patterns across communities, and this variability cannot be accounted for exclusively on the basis of linguistic variables, such as language typology and linguistic distance between contact languages  \cite{RN342,RN350}.  
Psycholinguists and neuroscientists are also increasingly interested in studying code-switching in experimental settings  \cite{RN360,RN544}, because it is then easier to control for a range of background variables. Computational approaches to code-switching have also gained momentum in the last decade \cite{solorio:2014,molina:2016,cetinoglu:2016c}.  The rise of social media and the ubiquitous presence of code-switching as found among multilingual online communities raised the profile of code-switching in the research agenda of computational linguistics and natural language processing. 

For researchers from all fields, it is crucially important to have access to code-switching corpora that have been carefully compiled and annotated, as these can provide the necessary information about the frequency of different types of code-switching in a particular language pair and social setting. 
While some code-switching corpora are already in the public domain (see Talkbank \citelanguageresource{RN51}, LinCE \citelanguageresource{aguilar:2020},  GLUECoS \citelanguageresource{khanuja:2020}),\footnote{\url{https://talkbank.org/}\\ \url{https://ritual.uh.edu/lince/datasets}
\url{https://microsoft.github.io/GLUECoS/}}\footnote{Bilingual corpora on the Talkbank include the following language pairs: English-Chinese; German-Greek; Greek-English, Hebrew-English, Japanese-Spanish, Spanish-English, Turkish-Danish, Welsh-Spanish; the LinCE corpus contains four language pairs: Spanish-English, Nepali-English, Hindi-English, and Modern Standard Arabic-Egyptian Arabic. Finally, the GLUECos comprises English-Hindi and English-Spanish data.}
there is a great need for more such corpora. 
This is because frequency patterns are more likely to be valid if they are drawn from a large database. 
Making data from a variety of sources accessible to the research community is also important because code-switching patterns can vary depending on the generation speakers belong to \cite{RN347}, or the type of bilinguals and the contexts they live in, such as, for example, heritage speakers of Turkish living in Germany versus returnees who live in Turkey \cite{RN354}. Less is known about differences between patterns found in oral conversations versus social media such as Twitter \citelanguageresource{RN371} or about the ways in which patterns can change over time. The greater the number of different corpora that are made available to the research community, the more likely it is that we can draw valid conclusions about the variability in code-switching patterns and the explanations for these patterns. This aim can only be achieved if corpora are presented in a computer-readable, standardised format, so that the data can be searched with computational tools.
Both labelled and unlabelled data are valuable in this regard to conduct experiments, evaluate results as well as computational aid for linguistic and psycholinguistic research questions.

The aim of the current paper is to provide information about the compilation of a Turkish-German bilingual corpus, called \tugebic. While some of the data have been analysed in \newcite{RN354,RN544} so far the entire corpus has not been available to the research community.
In this paper, we describe the process of collection, transcription, and annotation of the corpus. In addition, we provide token-level language IDs as automatic predictions, and describe quantitative and qualitative aspects of code-switching in the corpus.

\section{Data}
In this section we describe the collection of the corpus and give an overview of the transcription process that was initially carried out.

\subsection{Collection}
The data for the \tugebic corpus were collected in Germany and Turkey between 1993 and 1995 by a Turkish-German bilingual research assistant, who was born in Germany, but had returned to Turkey with his parents at the age of twelve. Free conversations were recorded with other undergraduate students at two different universities in Turkey as well as with master students from a university in Germany, and with family members and friends of the research assistant. Twelve males and 24 females were audio recorded for the study. They were between the ages of 18 and 50 years old, and the majority was in their early twenties. 

The audio recordings are no longer available and can therefore not be made available to the research community. 
Participants provided consent for being recorded, and for the analysis of the data for the purposes of the project.  
Permission to make the data available was obtained from the University of Reading, after consultation with the Ethics Committee and a data protection officer at this institution.  It has been agreed that data would only be donated after they had been carefully anonymised and sensitive passages had been deleted. No metadata will be released about the participants to protect their identities. 

All participants were Turkish-German bilinguals in that they had grown up with both languages and used them on a daily basis. No measurements of their knowledge of each language are available, but analyses of data from similar groups of Turkish-German bilinguals \cite{RN738,RN145} show that the Turkish language proficiency of the heritage speakers (that is speakers of Turkish who were born in  Germany and still live there) is lower than that of the returnees who had returned to Turkey in late adolescence and still live in Turkey. For German, the reverse is true. The same contrast can also be observed with respect to the amount of German and Turkish that is spoken by heritage speakers and returnees in the \tugebic corpus in that conversations the research assistant held with heritage speakers contain more German words than conversations held with returnees \cite{RN354}. 
The topics discussed during the recordings can be grouped into three main groups: a) students who discuss the content of classes and exams, as well as leisure activities. b) members of staff (two in total) who give a linguistics lecture; c) conversations among family members and friends of the research assistants, who talk about food, travel abroad, childhood experiences, leisure activities such as films they watched and sports they played, as well as differences between life in Turkey and Germany. Sensitive passages where participants discussed details of their private life were omitted.

\subsection{Transcription}
The transcription was carried out by the research assistant who was collecting the data, between 1993 and 1995. MS Word is chosen as the electronic format the transcripts are preserved. Here we outline the basic principles of the original transcription format.

\begin{itemize}
    \item \textbf{Metadata} All files had metadata, indicating time and place of recording as well as participants in the conversation and others who were present but did not contribute to the conversation.
    
    \item \textbf{Comments} The transcripts contained comments on the situation or on the start of a conversation.
    
    \item \textbf{Speaker ID} All speakers had a speaker ID, generally just a single capital letter, which represented the first letter of their first name, which was followed by a colon. Speaker turns were represented with this ID. If the identity of a speaker was unknown, they were represented as `X:' 
    \item \textbf{Multiple speakers} The speaker IDs were used in combination (e.g. A+B or A\&B) to indicate multiple speakers. Sometimes actual speaker names or kinship terms, such as \textit{Mum}, or generic descriptive terms, such as \textit{alle} ``all''were used.
    
    \item \textbf{Translations} In a few cases German translations of Turkish expressions were provided by the transcriber to make it more accessible to readers with higher German proficiency than Turkish.
    
    \item \textbf{Colloquial forms} Words were sometimes transcribed in colloquial forms, e.g., \textit{gidiyom} ``I am going" instead of the standard form \textit{gidiyorum}

    \item \textbf{Backchannels} Backchannels (brief verbal messages from listeners signaling that they are paying attention), such as \textit{mh} ``mm" or \textit{echt?} ``really?" etc., were put in parentheses in between the utterances of the main speaker in the original transcript.
    
    
    \item \textbf{Incomprehensible passages} Incomprehensible passages were marked with XXX (these were sometimes in upper and sometimes in lower case; the number of letters varied too). 
    
    \item \textbf{Pauses} Three or more dots were used to indicate pauses.
    
    \item \textbf{Hyphens} Between one and four hyphens are sometimes found in the middle of an utterance or at the end of an utterance, probably to indicate hesitations or incomplete sentences. A sequence of hyphens was used to indicate that a new conversation had started.
\end{itemize}



\section{Machine-Readability and Consistency Improvements}
In the 1990s, corpus data were rarely given to online databases, as these were still under development. To be able to publish the data, we revisited the transcriptions in 2021 and applied a series of manual and automatic improvements. A Turkish-German bilingual Masters student who studied Turcology at a German university worked as an annotator in the manual first stage, then semi-automatic corrections were made by a Computational Linguistics student (native German speaker) and the second author (native Turkish speaker) in the second stage.

\subsection{Manual Revisions}
\label{sec:phase1}
We formulated detailed guidelines to facilitate manual revisions. In this section we present the major points of these guidelines.

\paragraph{Format Conversion}
The very first step of the revision was moving the files from MS Word to .txt format to ensure machine-readibility and encoding consistency. All files were moved to UTF-8 encoding, and incorrect encodings of Turkish- and German-specific characters were replaced with the correct ones. At a later stage, we also unified the filenames and removed recurring content from files. The resulting collection contains 25 files.

\paragraph{Sentence Boundaries and Tokenisation}
Transcribing oral data is notoriously complex \cite{RN736}. There are no full stops in spoken language, for example, which makes it very difficult to determine where the boundary between sentences (or rather utterances) is. Whatever the direction of research one adopts, it remains important to determine where the boundaries between utterances are, which is never easy with oral data, because participants do not always speak in complete sentences: they may stop in the middle of an utterance or may decide to change the structure of an utterance during processing, for example, which means that rules for written language are not always observed in oral data.

For \tugebic we have adopted the choices of the transcriber, who indicated whether an utterance was finished by adding a full stop or another utterance terminator at the end of an utterance, albeit in the awareness that other transcribers might have made other choices.\footnote{As the recordings are no longer available it is not possible to ask for a second opinion on this issue.}
Whenever possible, we stayed as close as possible to the original. In this case an utterance could contain multiple sentences even they are separated by sentence-final punctuation. We nevertheless applied tokenisation, including adding spaces before punctuation marks excepts ones that are part of tokens (e.g., apostrophes in inflected Turkish proper names).

\paragraph{Intermediary Annotations} 
We introduced two intermediary annotation markers in the updated version of the corpus. These are
\begin{itemize}
    \item \textbf{$<$e$>$} for normalisation. For instance, \textit{sonar} is a typo of the Turkish word \textit{sonra} ``later". The annotator marked this correction as \texttt{sonar$<$e$>$sonra}.
    
    \item \textbf{$<$-$>$} for anonymisation. For instance, if \textit{Ali} is a name to be anonymised and the replacement is \textit{Veli}, the annotator indicated it in the corpus as \texttt{Ali$<$-$>$Veli}.
\end{itemize}

We chose annotated corrections instead of directly correcting them in the corpus for multiple reasons. When the annotator was in doubt they used a question mark as the replacement string. This way it was possible to easily reach all instances of unresolved cases. Moreover, the corrections are countable and reversible (e.g. for the debatable case of normalising colloquial use). The normalisation annotation could as well serve as a dataset for automatic normalisation tasks in the future.
In the second stage, the string before the annotation marker is replaced with the string after the marker automatically via a script.

\paragraph{Normalisation}
We corrected capitalisation where necessary, and inserted apostrophes before suffixes attached to proper names, as in \textit{{\.I}stanbul'da} ``in Istanbul". Restoring Turkish-specific characters, e.g. \texttt{bunlarin$<$e$>$bunların}, is also part of normalisation.
Non-standard orthography (e.g. \textit{gidiyom}) was replaced with standard orthography (\textit{gidiyorum} ``I am going"). 
We also corrected spelling errors and other typos, for example when an interrogative suffix was attached to the verb root, as in \textit{biliyormusun} ``do you know". According to standard spelling rules the question suffix needs to be separated as in \textit{biliyor musun}. The same type of corrections were applied to writing \textit{de} and \textit{ki} clitics, which are common mistakes also in Turkish social media.

The remaining revision steps could be summarised as follows:
\begin{itemize}
    \item \textbf{Intra-word Code-switching} When a token contains morphemes from two languages, we call it intra-word code-switching. To facilitate the study of such mixed tokens, these were marked with a paragraph mark §\footnote{This sign is chosen by \citelanguageresource{RN371} for practical purposes, that is, the Twitter corpus that contains a diverse list of characters did not contain this particular character so that it could be used as a unique identifier. As our corpus also did not contain this character we follow the existing convention.} at the boundary between the two languages, as in \textit{Stadt§tan} (city§\textsc{abl} meaning ``from the city"), where § indicates the boundary between the German root \textit{Stadt} ``city" and the Turkish ablative suffix \textit{-tan}. 

    
    \item \textbf{Pauses} Where in the original transcript three or more dots were used to indicate pauses, these were replaced with the string \texttt{[pause]}. 
    
    
    \item \textbf{Backchannels} Backchannels were put on separate lines and provided with a speaker ID.
    
    \item \textbf{Unintelligible material} The various representations of XXX were replaced with the annotation \texttt{[unintelligible]}.
    
    \item \textbf{Translations} Translations were omitted from the final transcript, as translations were not given for complete sentences, and creating glosses for the entire corpus was beyond the scope our project.
\end{itemize}

\subsection{Sanity and Consistency Checks}
An automatic check was done to verify the manual revisions of the first phase and to find out remaining mistakes. The corrections that need decisions are resolved manually, the remaining ones are addressed automatically. This phase can be summarised as follows:

\begin{itemize}
    \item During manual revision we asked the annotator to use a question mark as the replacement string when they are uncertain about a normalisation or anonymisation case. 
    All such cases were resolved with appropriate replacements.
    
    \item In the original files there were a few instances of \textit{X} used instead of person names in utterances (e.g. \textit{X war doch schon verheiratet} ``X was already married"). They were replaced with the string \texttt{[anon]} so that they can be distinguished from the speaker ID \textit{X}.
    
    \item Remaining human errors were corrected such as annotation mistakes (e.g. typo in \texttt{[unintelligible]}), tokenisation, unmatched parentheses, idiosyncratic characters.
    
    \item Replacements for the intermediate anonymisation and normalisation annotations were carried out.
    
    \item Speaker IDs were standardised to a single character for a speaker followed by a column sign, and there is a space before the utterance starts.  
    
    \item The representation of multiple speakers were unified. Two or more speakers were represented their respective single characters combined with a `\&' sign.
    
    \item The hyphens in the middle of a sentence (between one and three) were left in place. The number of the hyphens at the end was standardised to three. Four or more hyphens on an otherwise empty line were converted to twenty `=', denoting conversation separators.
    
    \item  Empty lines, trailing spaces were removed, and multiple spaces between tokens were reduced to one. 
\end{itemize}

\subsection{Anonymisation}
All names of speakers were replaced with names which started with the same letter as the original name, and ensured vowel harmony rules would remain the same for the pseudonym. Thus, for example, \textit{Gülsen} was chosen as the pseudonym for a female name which started with a G and for which the last syllable contained the front vowel ``e". For names mentioned in the transcripts (not participants in the project) we kept the format the original transcripts had for names. When the name was fully transcribed (e.g. \textit{Ali}), we anonymised it, but when only the first letter was transcribed  (e.g. A.) we did not anonymise the names further. We did not anonymise celebrities, e.g. \textit{Mustafa Sandal}, a Turkish pop singer either. 

Place names were replaced with alternative place names which sounded similar in that they contained the vowels with the same articulatory properties in the last syllable as the original, so that any suffixes on the names could be maintained and vowel harmony rules were respected.

Furthermore, words and phrases that could reveal the identity of the speaker due to context were also be anonymised. For instance, a course name was anonymised in transcriptions where students talked about the degree course they were taking. Finally, passages which revealed sensitive personal information about the speaker(s) were deleted.

All anonymisation is done manually. In the sanity check phase we extracted the list of anonymisation pairs and automatically checked if any references in the list were left out. We then corrected the mistakes or kept the proper name as is depending on the context (e.g. the Turkish female name \textit{Ay\c{s}e} is not a sensitive reference when used in \textit{Ay\c{s}e fasulye}, a type of string beans). 

\subsection{Language ID Prediction}
\label{sec:lid}
The original version of the corpus is not annotated with language IDs. Instead, we provide predicted language IDs for each token. We trained the STEPS tool \citelanguageresource{grunewald:2021} that utilises XLM-RoBERTa\citelanguageresource{conneau:2020} as a language identifier.

As training data we used the 2.8 version of the Turkish-German SAGT Treebank \citelanguageresource{cetinoglu:2019}.\footnote{\texttt{github.com/UniversalDependencies/}\\\texttt{UD\_Turkish\_German-SAGT}} The size of the training data in the standard split is quite small (578 sentences), which might affect the performance in a new dataset. Since our intention is to annotate the \tugebic corpus as accurately as possible, we increased the training data size by combining the standard training and test splits. The model performs with 98.8\% accuracy on the development set of the treebank. As there are no gold language IDs on the \tugebic corpus, it is not possible to evaluate the accuracy of the model on this dataset. Nevertheless the in-domain accuracy of the language identifier is quite high and both \tugebic and SAGT are transcriptions of bilingual speaker conversations, hence it is likely that the same model can perform well on the new dataset too.

The SAGT Treebank employs the following language ID tagset: 
\begin{itemize} 
\setlength{\itemsep}{-1pt}
    \item \texttt{TR} for Turkish
    \item \texttt{DE} for German
    \item \texttt{LANG3} for tokens that belong to a third language other than Turkish and German
    \item \texttt{MIXED} for tokens with intra-word code-switching    
    \item \texttt{OTHER} for punctuation, symbols, and any other tokens that do not fall into the categories above
\end{itemize}
 
Before predicting language IDs, we did automatic sentence segmentation. We used dot, exclamation mark, and question mark as sentence delimiters if they are separate tokens in the middle of a line. If the sentence-final punctuation was inside quotation marks, we assumed that they are part of a larger sentence, e.g., reported speech, and did not apply the split. If segmentation resulted in a sentence consisting only of \texttt{[pause]} token followed by a sentence-final punctuation, we removed this sentence. We also removed all \texttt{[pause]} tokens, speaker IDs and conversation separators.

After predicting language IDs we automatically applied minor postprocessing to the output based on heuristics. We made sure that all \texttt{[unintelligible]} and \texttt{[anon]} tokens, and all punctuation are assigned the \texttt{OTHER} label, and all intra-word code-switching tokens with the \S{} annotation  have the \texttt{MIXED} language ID.

\section{Corpus Characteristics}
In this section we give a quantitative and qualitative analysis of the corpus. The quantitative analysis includes the basic statistics of the corpus as well as language ID and code-switching point distributions according to the predicted language IDs. Although the statistics based on predictions are not 100\% correct, we believe they are useful instruments to portray the overview of the corpus with a close approximation of the real distribution.

As qualitative analysis we present a set of observations of code-switching patterns from the corpus. In all examples German words are shown in bold font and Turkish words in regular font. A morpheme-by-morpheme analysis is given when the interpretation of the examples required it.\footnote{The abbreviations used in the examples: 1PL = first person plural; 1SG = first person singular; 3SG = third person singular; AOR = aorist; COND = conditional; DAT = dative; GEN = genitive; LOC = locative; NMLZ = nominalization; PASS = passive, PL = plural; PROG = progressive; PST = past; PTCP = participle; REL = relative.} In providing these interlinear glosses, we follow the Leipzig Glossing rules \cite{RN743}. The numbered `Tugebic' string that is presented in parentheses after the translation denotes the file the example is taken from.

\subsection{Statistics}
\begin{table}[t!]
 \setlength{\tabcolsep}{-4pt}
 \centering
 \begin{tabular}{l S[table-format=6.2] r}
 \hline
 \multicolumn{2}{l}{\textit{Basics}}\\
 \hline
\# of tokens &  116688 & \\
\# of sentences & 14651 & \\
\# of monolingual sentences & 10141\\
\# of bilingual sentences & 4510\\
\# of CS points in bilingual sentences&   8180 & \\
Avg sentence length & 7.96 & \\
Avg CS points per bilingual sentence & 1.82  \\
\hline 
\multicolumn{2}{l}{\textit{Language ID Distribution}}\\
\hline
\texttt{TR}       &  43785 & (37.52\%)\\
\texttt{DE}        & 43210 & (37.03\%)\\
\texttt{OTHER}     & 28656 & (24.56\%)\\
\texttt{MIXED}  &     686 & (0.59\%)\\
\texttt{LANG3}    &    351 & (0.30\%)\\
\hline
 \end{tabular}
 
 \caption{The basic statistics of the \tugebic corpus and the language ID distribution according to predicted language IDs. CS denotes code-switching.}
 \label{tab:stats}
\end{table}


The basic statistics on the corpus and the language ID distribution are given in Table \ref{tab:stats}. The number of sentences is based on the automatic segmentation. The minimum sentence length is 1 and the maximum is 202 tokens. The majority of the corpus consists of monolingual sentences.
Only the sentences that contain intrasentential or intra-word code-switching (i.e., \textit{bilingual sentences}) are considered in calculating the code-switching points.\footnote{For instance, there are three code-switching points in (\ref{ex:Einkaufsstr}) and two code-switching points in (\ref{ex:zayif}).}
There are a total of 8180 code-switching points, which correspond to an average of 1.82 points per bilingual sentence. This figure is in line with the Turkish-German Twitter Corpus \citelanguageresource{RN371} and the SAGT Corpus \citelanguageresource{cetinoglu:2019} that both consist of only bilingual sentences and that both have averages slightly over 2 points. 

In the language ID distribution, however, these corpora show different behaviour. In the Twitter corpus \texttt{TR} tokens are dominant, and in the SAGT Corpus, the \texttt{DE} tokens take approximately half of the corpus. In our case the Turkish and German tokens are almost balanced. The differences are likely due to the fact that the \tugebic corpus also contains monolingual sentences, and not just a selection of utterances with code-switching.

5565 of the monolingual sentences are in Turkish and 4323 are in German. They are followed by 1582 sentences that start in German and continue in Turkish, and for 853 sentences it is the other way around. They constitute the majority of the sentences with a single code-switching point (the rest are sentences with the combinations of \texttt{LANG3} with \texttt{TR} or \texttt{DE}). 1246 sentences contain two switches, 329 sentences contain three switches, and 216 sentences contain four switches. Sentences with five or more switches are 234 in total. The highest number of switches is 20, with only one instance. 

\subsection{Observations}

Transcriptions contain a large number of switches between utterances (\textit{intersentential code-switching}), as in (\ref{ex:beiuns}), where Speaker K makes a statement in German, and Speaker O replies in German, after which Speaker K answers in Turkish.\footnote{One reviewer suggests it would have been preferable to give exact numbers of different types of switches. This is only possible if the corpus is manually annotated with switch points and ideally with syntactic structures. Thus this is beyond the scope of the current paper.}

\begin{exe}
 \ex \label{ex:beiuns}
K: \textbf{Bei uns kannst du nicht so ohne Jacke}. [pause].\\
O: \textbf{Warum}?\\
K: Buz gibi.
 \trans `K: \textbf{At our place you cannot go without a jacket}. O: \textbf{Why}? K: Cold as ice.' (21Tugebic)
\end{exe}

While in (\ref{ex:beiuns}) the switch takes place between interlocutors, intersentential code-switching can also take place between utterances of the same speaker, as in (\ref{ex:angefangen}), where Speaker S switches to Turkish after her first utterance in German.\footnote{Note that we take proper names as language-specific, hence they take part in code-switching as exemplified with \textit{Seçil} in (\ref{ex:beiuns}). While many proper names keep their original form across languages, there are also many language-specific ones as in \textit{Munich} in English, \textit{München} in German, and \textit{Münih} in Turkish. And even if the proper name does not change, it is still an unknown word for the other language in many cases, e.g., there will be no instances of \textit{Seçil} in monolingual resources of German.}

\begin{exe}
 \ex \label{ex:angefangen} 
A: \textbf{Und wie haben wir angefangen}, Seçil? \\
S: \textbf{Na ja, weiß ich nicht}. Yok bunun babası ilk defa beni şeye bıraktı. 
 \trans `A: \textbf{And how did we start}, Seçil? S: \textbf{Well, I don't know}. No, his father first of all left me at thingie.' (21Tugebic)
\end{exe}

Switching within utterances (intrasentential code-switching) takes place very frequently in the dataset. For the current purposes, we use the four-way typology of  \cite{RN350} to describe the different types of code-switching in the data. One of the most frequent types of switching in the data is \textsc{insertion}: informants often insert a content word or fixed expression from German into Turkish as in (\ref{ex:Einkaufsstr}), where the German compound \textit{Einkaufsstraße} ``shopping road" is inserted into a Turkish sentence, or vice versa, as in (\ref{ex:zayif}), where a Turkish noun \textit{zayıf} ``low mark" is inserted into a German sentence.

\begin{exe}
 \ex \label{ex:Einkaufsstr}
 Böyle \textbf{Einkaufsstraße} gibi bir şey var, \textbf{weißt du in Darmstadt}! 
 \trans`There is a kind of \textbf{shopping street} or something there, \textbf{you know, in Darmstadt}.' (19Tugebic, Speaker A)
\end{exe}


\begin{exe}
 \ex \label{ex:zayif}
\textbf{Es gibt keine} zayıf§\textbf{s}.
\trans`\textbf{There are no} low mark\textbf{s}.' (20Tugebic, Speaker M).
\end{exe}

In (\ref{ex:Einkaufsstr}) and (\ref{ex:zayif}) there is a clear matrix language \cite{myers1997duelling}, which sets the grammatical frame. In (\ref{ex:Einkaufsstr}) the matrix language is Turkish, and in (\ref{ex:zayif}) it is German. 
When the matrix language is Turkish, German nouns that are inserted into Turkish can receive Turkish inflection, as in  (\ref{ex:mittlerweile}), where German \textit{Wohnheim} receives the Turkish plural \textit{-lar}, which is followed by the Turkish genitive \textit{-ın}.

\begin{exe}
 \ex \label{ex:mittlerweile}
 \gll \textbf{Mittlerweile} bu şey-ler-in \textbf{Wohnheim}§-lar-ın hepsi bir sistem-e bağla-n-dı. \\
   \textbf{{in the meantime}} that thing-PL-GEN \textbf{dormitory}§-PL-GEN all one system-DAT link-PASS-PST.\\
 \trans `\textbf{In the meantime}, those things, \textbf{dormitori}es,  were all connected to one system.' (14Tugebic, speaker S) 
\end{exe}

It is interesting that vowel harmony rules apply across language boundaries. In (\ref{ex:mittlerweile}), for example, the second syllable of \textit{Wohnheim} ``dormitory" contains the diphthong \textipa{/aI/},
which means that a plural suffix with the back vowel \textit{a}, namely \textit{-lar}, needs to be chosen to mark this noun for number.
The opposite (German inflection on Turkish nouns) can be seen in \ref{ex:zayif}, where \textit{zayıf} ``low (mark)" receives a German \textit{-s} plural, although this is much rarer in the data than Turkish marking of plurals on German nouns.

Another fairly rare type of code-switch can be found in (\ref{ex:Zeitung}). Here \textit{die Zeitung} ``the newspaper" is probably the subject of the verb \textit{ilan ediyor} ``advertises", although this switch point is untypical because switching between the subject and the verb is severely constrained in most language combinations \cite{RN845}. Another reason why this example is exceptional is that in Turkish-German code-switching data we rarely find switches of full Determiner Phrases, such as the one in \ref{ex:Zeitung}: the German determiner \textit{die} "the" accompanies the German noun \textit{Zeitung} ``newspaper".

\begin{exe}
 \ex \label{ex:Zeitung}
 \gll Gazete-de ilan aaa \textbf{die Zeitung} ilan, ilan ed-iyor.\\
 Newspaper-LOC advertise ehm \textbf{the newspaper} advertise, advertise do-PROG\\
 \trans`He advertises in the newspaper ehm,  \textbf{the newspaper} advertises (it).' (18Tugebic, Speaker Ö)
\end{exe}

While informants frequently produce insertions such as those in \ref{ex:Einkaufsstr}, switches can also occur at the periphery of a sentence, for example for an adverbial phrase, such as \textit{mittlerweile} ``in the mean time" in (\ref{ex:mittlerweile}). This type of switching is called \textsc{alternation}. Alternations are often longer and can, for example, consist of an entire adverbial clause, as in (\ref{ex:buraya}), which begins with a Turkish adverbial clause marked with \textit{-ken} ``when/while", whereas the main clause \textit{hab' ich mich gefreut} ``I was happy" is in German. It is also remarkable that the Turkish adverbial clause can trigger Verb Second in German, as can be seen in the occurrence of the inflected verb \textit{hab'} ``have" directly after the adverbial clause in (\ref{ex:buraya}).

\begin{exe}
\ex \label{ex:buraya}
\gll Ben bura-ya gel-ir-ken \textbf{hab'} \textbf{ich} \textbf{mich} \textbf{gefreut}.\\
 I here-DAT come-AOR-when \textbf{have} \textbf{I} \textbf{myself} PTCP-\textbf{be.happy}-PTCP\\
\trans `When I came here, \textbf{I was happy}.' (03Tugebic, Speaker H)
\end{exe}

A special case of alternation can be found in (\ref{ex:wenn}), where the switch takes place in the middle of a conditional clause: the German conditional subordinate conjunction \textit{wenn} ``if" is used at the start of the sentence, followed by the German pronoun \textit{wir} ``we" but the inflected verb \textit{olsaydık} also contains the information that this is a conditional clause, in the suffix \textit{sa} ``if" and it is also inflected for person. The doubling of grammatical information in German and Turkish is typical for alternational types of switches \cite{RN350,cetinoglu:2019}.

\begin{exe}
 \ex \label{ex:wenn}
 \gll \textbf{Wenn wir}  yakın ol-say-dı-k. \\
 \textbf{If we} close be-COND-PST-1PL\\
\trans`\textbf{If we} were close.' (01Tugebic, Speaker U)
\end{exe}

The third type, \textsc{congruent lexicalisation}, is one where both German and Turkish contribute grammatical and lexical items and the structure becomes a hybrid between both languages, as in (\ref{ex:okutman}), where a German copula \textit{ist} ``is" links the subject \textit{okutman} ``teacher" and the subject complement \textit{öǧretmen} ``teacher".

\begin{exe}
 \ex \label{ex:okutman}
 \gll Okutman \textbf{ist} sonuç-ta öǧretmen yani, normal öǧretmen \textbf{ohne} \textbf{Dings}. \\
 Lecturer \textbf{ist} end-LOC teacher {you know}, normal teacher \textbf{without} \textbf{thingie}.\\
 \trans`A lecturer \textbf{is} a teacher after all, you know, a normal teacher without a \textbf{what's it called again}.' (03Tugebic, Speaker K)
\end{exe}
 
The fourth type of code-switching, called \textsc{backflagging}, involves the use of a discourse marker from the heritage language (Turkish) in discourse that is otherwise entirely in German. This is called  backflagging because this strategy can be used to flag up one's ethnic identity \cite[p. 713]{RN350}, as in (\ref{ex:iste}), which begins with the Turkish discourse marker \textit{işte} ``well."

\begin{exe}
 \ex \label{ex:iste}
 İşte, \textbf{ich bin ausgestiegen}.
 \trans`Well, \textbf{I got off} (the bus).' (19Tugebic, Speaker A)
\end{exe}

Backflagging appears to be particularly frequent among the Germany-based speakers in the \tugebic corpus possibly because they mainly speak more German than Turkish in daily life. 

Another very frequent type of code-switching is one where the Turkish verb \textit{yap-} ``to do/make" is used as a light verb and combined with a German verb in the infinitive form, as in (\ref{ex:yap}), where German \textit{anmelden} ``sign in" and \textit{auflisten} ``to list" are used in combination with the Turkish light verb \textit{yap-} ``to do/make".

\begin{exe}
 \ex \label{ex:yap}
 \gll Ora-da bekle-me şey-i var, liste-si var, işte \textbf{anmelden} yap-an-lar-ın hepsi, ora-ya  \textbf{auflisten} yap-ıl-ıyor.\\
 There-LOC wait-NMLZ thing-POSS {there is}, list-POSS, {there is} there \textbf{sign.in} do-REL-PL-GEN all, there-DAT \textbf{list} do-PASS-PROG.3SG\\
 \trans `There was a waiting thing, a list, there were  people \textbf{signing in} and \textbf{lists were made}.' (14Tugebic, Speaker S)
\end{exe}
 
In examples with mixed verbal compounds, the matrix verb \textit{yap-} carries the inflection, and the embedded German verb is in the infinitive form. This type of bilingual compound verb construction is found in many Turkish-speaking immigrant communities in Europe, e.g. in the Netherlands \cite{RN347}, Germany \cite{RN356,RN360}\citelanguageresource{RN371,cetinoglu:2019} or Norway \cite{RN739}. According to \cite{RN342}, these forms can be seen as insertions or alternations, depending on the type of element that is embedded in this construction, and the relationship between the embedded verb and the matrix verb. Interestingly, in the \tugebic corpus, one of the teachers avoided the light verb construction using \textit{yap-}, and employed \textit{et-} ``do/make" as the matrix verb instead to create a mixed verbal compound, as in (\ref{ex:ediyorlar}). This form is frequently used in standard Turkish nominal verb compounds \cite{RN740}, but less common in mixed verbal compounds as found among heritage speakers. 

\begin{exe}
 \ex \label{ex:ediyorlar}
 \gll Hatta \textbf{widersprechen} ed-iyor-lar.\\
 {In fact} \textbf{contradict} do-PROG-PL\\
 \trans`In fact, they \textbf{contradict} (that).' (02Tugebic, Speaker G)
\end{exe}

The tutor may have preferred \textit{et-} because compound verbs with \textit{yap-} are typically used by heritage speakers of Turkish but less often by Turkish people who have always lived in Turkey \cite{RN360,RN356}, such as the tutor who produced this utterance. 

Finally, there were a few exceptional cases of switches within words, as in \ref{ex:kommiyom}, where Turkish tense \textit{-iyo} and person marking \textit{-m} is attached to the German root \textit{komm} ``come". This is a quite surprising use as the preferred way in such constructions is employing light verbs as exemplified in  (\ref{ex:yap}) and (\ref{ex:ediyorlar}). 

\begin{exe}
 \ex \label{ex:kommiyom}
 \gll \textbf{Ich} \textbf{komm}-iyo-m$<$e$>$\textbf{komm}§iyorum \\
 \textbf{I} \textbf{come}-PROG-1SG\\
 \trans`I am \textbf{com}ing.' (01Tugebic, Speaker U)
\end{exe}

The other rare type of intra-word switch is (\ref{ex:compound}), where a Turkish noun \textit{kayın} ``in-law" and a German noun \textit{Sohn} ``son" are combined into a mixed compound. The speaker probably meant \textit{son-in-law} as they repeated the same word in full German as the next token. There are some kinship terms in Turkish that start with \textit{kayın}, e.g., \textit{kayınpeder} ``father-in-law", \textit{kayınvalide} ``mother-in-law", \textit{kayınbirader} ``brother-in-law" but son-in-law is not one of them. It seems the speaker notices this mid-way and switches to German. 


\begin{exe}
 \ex \label{ex:compound}
 \gll Almanya'da ben-im kayın§\textbf{sohn}, Schwieger-sohn\\
 Germany-LOC I-1SG.POSS {in-law}§\textbf{son}, {in-law-son}.\\
 \trans`My \textbf{son}-in-law is in Germany." (01Tugebic, Speaker A)
\end{exe}

Further research on the corpus will need to be carried out to determine the frequency of the different code-switching types in the corpus, and the sociolinguistic variables associated with this variability. 

\section{Conclusion}
In this paper, we have described the characteristics of a corpus of spontaneous conversations among Turkish-German bilinguals in Turkey and in Germany called \tugebic. We have have summarised the steps taken for the data collection, and given an overview of the transcription and further data treatment of the corpus. Furthermore, we have presented quantitative details about the number of monolingual and bilingual sentences and provided examples of different types of code-switching in the data. 
To the best of our knowledge, this is the largest Turkish-German dataset that is made available to the research community to date.

As the \tugebic data were collected in the 1990s, they offer an historical perspective on code-switching as it was common in this bilingual speech community about 30 years ago. A possible avenue for future research could be to compare this dataset to more recent datasets  \citelanguageresource{RN352,cetinoglu:2019} to establish whether code-switching patterns found currently differ from those in the 1990s, or to compare code-switching patterns from spontaneous conversations with those found in social media, such as Twitter \citelanguageresource{RN371}. 

The corpus is also valuable as an additional resource in computational studies, especially given that no other unlabelled Turkish-German code-switching dataset is available. Both the standard corpus files and the files with predicted language IDs can be obtained from 
\url{https://github.com/ozlemcek/TuGeBiC}.

\section{Acknowledgements}
We are very grateful to Kubilay Yalçın who collected and transcribed the data, to Mustafa Cem Güneş who tokenised and normalised the data, and to Marina Haid who implemented automatic sanity checks and corrections, and ran language ID experiments. The first author also gratefully acknowledges sponsoring from the Language Contact Fund of the University of Amsterdam for the data collection. 
The second author is funded by DFG via project CE 326/11 \textit{Computational Structural Analysis of German Turkish Code-Switching (SAGT)}.
This paper was accepted for LREC 2022 but subsequently withdrawn. We are thankful to the three anonymous reviewers for their helpful comments.

\section{Bibliographical References}\label{reference}

\bibliographystyle{lrec2022-bib}
\bibliography{lrec2022-example}

\begin{thebibliography}{}

\bibitem[\protect\citename{Aguilar \bgroup et al.\egroup }2020]{aguilar:2020}
Aguilar, G., Kar, S., and Solorio, T.
\newblock (2020).
\newblock {L}in{CE}: {A} {C}entralized {B}enchmark for {L}inguistic
  {C}ode-switching {E}valuation.
\newblock In {\em Proceedings of The 12th Language Resources and Evaluation
  Conference}, pages 1803--1813, Marseille, France, May. European Language
  Resources Association.

\bibitem[\protect\citename{\c{C}etino{\u{g}}lu}2016]{RN371}
\c{C}etino{\u{g}}lu, {\"O}.
\newblock (2016).
\newblock A {Turkish-German} code-switching corpus.
\newblock In Nicoletta Calzolari, editor, {\em Proceedings of the Tenth
  International Conference on Language Resources and Evaluation}, pages
  4215--4220. European Language Resources Association.

\bibitem[\protect\citename{{\c{C}}etino{\u{g}}lu and
  {\c{C}}{\"o}ltekin}2019]{cetinoglu:2019}
{\c{C}}etino{\u{g}}lu, {\"O}. and {\c{C}}{\"o}ltekin, {\c{C}}.
\newblock (2019).
\newblock Challenges of annotating a code-switching treebank.
\newblock In {\em Proceedings of the 18th International Workshop on Treebanks
  and Linguistic Theories (TLT, SyntaxFest 2019)}, pages 82--90, Paris, France,
  August. Association for Computational Linguistics.

\bibitem[\protect\citename{Conneau \bgroup et al.\egroup }2020]{conneau:2020}
Conneau, A., Khandelwal, K., Goyal, N., Chaudhary, V., Wenzek, G., Guzm{\'a}n,
  F., Grave, E., Ott, M., Zettlemoyer, L., and Stoyanov, V.
\newblock (2020).
\newblock Unsupervised cross-lingual representation learning at scale.
\newblock In {\em Proceedings of the 58th Annual Meeting of the Association for
  Computational Linguistics}, pages 8440--8451, Online, July. Association for
  Computational Linguistics.

\bibitem[\protect\citename{Doğruöz and Backus}2009]{RN352}
Doğruöz, A.~S. and Backus, A.
\newblock (2009).
\newblock Innovative constructions in dutch turkish: An assessment of ongoing
  contact-induced change.
\newblock {\em Bilingualism: language and cognition}, 12(01):41--63.

\bibitem[\protect\citename{Gr{\"u}newald \bgroup et al.\egroup
  }2021]{grunewald:2021}
Gr{\"u}newald, S., Friedrich, A., and Kuhn, J.
\newblock (2021).
\newblock Applying occam{'}s razor to transformer-based dependency parsing:
  What works, what doesn{'}t, and what is really necessary.
\newblock In {\em Proceedings of the 17th International Conference on Parsing
  Technologies and the IWPT 2021 Shared Task on Parsing into Enhanced Universal
  Dependencies (IWPT 2021)}, pages 131--144, Online, August. Association for
  Computational Linguistics.

\bibitem[\protect\citename{Khanuja \bgroup et al.\egroup }2020]{khanuja:2020}
Khanuja, S., Dandapat, S., Srinivasan, A., Sitaram, S., and Choudhury, M.
\newblock (2020).
\newblock {GLUEC}o{S}: An evaluation benchmark for code-switched {NLP}.
\newblock In {\em Proceedings of the 58th Annual Meeting of the Association for
  Computational Linguistics}, pages 3575--3585, Online, July. Association for
  Computational Linguistics.

\bibitem[\protect\citename{MacWhinney}2000]{RN51}
MacWhinney, B.
\newblock (2000).
\newblock The {CHILDES} project: Tools for analyzing talk: Volume i:
  Transcription format and programs, volume ii: The database.
\newblock {\em Computational Linguistics}, 26(4):657--657.

\end{thebibliography}


\begin{thebibliography}{}

\bibitem[\protect\citename{Backus}1996]{RN347}
Backus, A.
\newblock (1996).
\newblock {\em Two in one. Bilingual speech of Turkish immigrants in the
  Netherlands}.
\newblock Studies in multilingualism. Tilburg University Press, Tilburg.

\bibitem[\protect\citename{{\c{C}}etinoglu \bgroup et al.\egroup
  }2016]{cetinoglu:2016c}
{\c{C}}etinoglu, {\"{O}}., Schulz, S., and Vu, N.~T.
\newblock (2016).
\newblock Challenges of computational processing of code-switching.
\newblock In Mona~T. Diab, et~al., editors, {\em Proceedings of the Second
  Workshop on Computational Approaches to Code Switching@EMNLP 2016, Austin,
  Texas, USA, November 1, 2016}, pages 1--11. Association for Computational
  Linguistics.

\bibitem[\protect\citename{Comrie \bgroup et al.\egroup }2015]{RN743}
Comrie, B., Haspelmath, M., and Bickel, B.
\newblock (2015).
\newblock The leipzig glossing rules: Conventions for interlinear
  morpheme-by-morpheme glosses.
\newblock Report, Department of Linguistics of the Max Planck Institute for
  Evolutionary Anthropology and the Department of Linguistics of the University
  of Leipzig.

\bibitem[\protect\citename{Creese and
  Blackledge}2010]{creese2010translanguaging}
Creese, A. and Blackledge, A.
\newblock (2010).
\newblock Translanguaging in the bilingual classroom: A pedagogy for learning
  and teaching?
\newblock {\em The modern language journal}, 94(1):103--115.

\bibitem[\protect\citename{Daller \bgroup et al.\egroup }2011]{RN145}
Daller, M.~H., Yıldız, C., de~Jong, N.~H., Kan, S., and Başbağı, R.
\newblock (2011).
\newblock Language dominance in turkish–german bilinguals: Methodological
  aspects of measurements in structurally different languages.
\newblock {\em International Journal of Bilingualism}, pages 215--236.

\bibitem[\protect\citename{Daller}1996]{RN738}
Daller, H.
\newblock (1996).
\newblock {\em Migration und Mehrsprachigkeit. Der Sprachstand türkischer
  Rückkehrer aus Deutschland}.
\newblock Peter Lang, Frankfurt am Main.

\bibitem[\protect\citename{Kerslake and Goksel}2014]{RN740}
Kerslake, C. and Goksel, A.
\newblock (2014).
\newblock {\em Turkish: an essential grammar}.
\newblock Routledge, London.

\bibitem[\protect\citename{MacSwan}2000]{RN845}
MacSwan, J.
\newblock (2000).
\newblock The architecture of the bilingual language faculty: Evidence from
  intrasentential code switching.
\newblock {\em Bilingualism: language and cognition}, 3(1):37--54.

\bibitem[\protect\citename{MacWhinney}2019]{RN736}
MacWhinney, B.
\newblock (2019).
\newblock Understanding spoken language through talkbank.
\newblock {\em Behavior research methods}, 51(4):1919--1927.

\bibitem[\protect\citename{Molina \bgroup et al.\egroup }2016]{molina:2016}
Molina, G., AlGhamdi, F., Ghoneim, M., Hawwari, A., Rey-Villamizar, N., Diab,
  M., and Solorio, T.
\newblock (2016).
\newblock Overview for the second shared task on language identification in
  code-switched data.
\newblock In {\em Proceedings of the Second Workshop on Computational
  Approaches to Code Switching}, pages 40--49, Austin, Texas, November.
  Association for Computational Linguistics.

\bibitem[\protect\citename{Muysken}2000]{RN342}
Muysken, P.
\newblock (2000).
\newblock {\em Bilingual speech: A typology of code-mixing}, volume~11.
\newblock Cambridge University Press.

\bibitem[\protect\citename{Muysken}2013]{RN350}
Muysken, P.
\newblock (2013).
\newblock Language contact outcomes as the result of bilingual optimization
  strategies.
\newblock {\em Bilingualism: Language and cognition}, 16(04):709--730.

\bibitem[\protect\citename{Myers-Scotton}1997]{myers1997duelling}
Myers-Scotton, C.
\newblock (1997).
\newblock {\em Duelling languages: Grammatical structure in codeswitching}.
\newblock Oxford University Press.

\bibitem[\protect\citename{Pfaff}2000]{RN360}
Pfaff, C.~W., (2000).
\newblock {\em Development and use of et-and yap-by Turkish/German bilingual
  children}, pages 365--373.
\newblock Harrassowitz, Wiesbaden.

\bibitem[\protect\citename{Solorio \bgroup et al.\egroup }2014]{solorio:2014}
Solorio, T., Blair, E., Maharjan, S., Bethard, S., Diab, M., Ghoneim, M.,
  Hawwari, A., AlGhamdi, F., Hirschberg, J., Chang, A., and Fung, P.
\newblock (2014).
\newblock Overview for the first shared task on language identification in
  code-switched data.
\newblock In {\em Proceedings of the First Workshop on Computational Approaches
  to Code Switching}, pages 62--72, Doha, Qatar, October. Association for
  Computational Linguistics.

\bibitem[\protect\citename{Treffers-Daller \bgroup et al.\egroup }2016]{RN356}
Treffers-Daller, J., Daller, M., Furman, R., and Rothman, J.
\newblock (2016).
\newblock Ultimate attainment in the use of collocations among heritage
  speakers of turkish in germany and turkish–german returnees.
\newblock {\em Bilingualism: Language and Cognition}, 19(03):504--519.

\bibitem[\protect\citename{Treffers-Daller}1997]{RN354}
Treffers-Daller, J., (1997).
\newblock {\em Variability in codeswitching styles: Turkish-German
  code-switching patterns}, volume~1, pages 277--297.
\newblock Mouton de Gruyter, Berlin.

\bibitem[\protect\citename{Treffers-Daller}2020]{RN544}
Treffers-Daller, J., (2020).
\newblock {\em Turkish-German code-switching patterns revisited: what
  naturalistic data can(not) tell us.}
\newblock John Benjamins, Amsterdam/Philadelphia.

\bibitem[\protect\citename{Türker}2000]{RN739}
Türker, E.
\newblock (2000).
\newblock {\em Turkish-Norwegian codeswitching: Evidence from intermediate and
  second generation Turkish immigrants in Norway}.
\newblock Thesis.

\end{thebibliography}

\section{Language Resource References}
\label{lr:ref}
\bibliographystylelanguageresource{lrec2022-bib}
\bibliographylanguageresource{languageresource}

\end{document}